\documentclass[10pt,twocolumn,letterpaper]{article}

\usepackage{cvpr}              

\usepackage{graphicx}
\usepackage{amsmath}
\usepackage{amssymb}
\usepackage{booktabs}
\usepackage{pifont}
\usepackage{multirow}
\usepackage{wrapfig}
\usepackage{rotating}
\usepackage{xcolor}
\usepackage[accsupp]{axessibility}  

\newcommand{\ours}{OVSeg }

\definecolor{ForestGreen}{RGB}{34,139,34}

\usepackage[pagebackref,breaklinks,colorlinks]{hyperref}

\usepackage[capitalize]{cleveref}
\crefname{section}{Sec.}{Secs.}
\Crefname{section}{Section}{Sections}
\Crefname{table}{Table}{Tables}
\crefname{table}{Tab.}{Tabs.}

\def\cvprPaperID{8751} 
\def\confName{CVPR}
\def\confYear{2023}

\begin{document}

\title{\vspace{-1em}Open-Vocabulary Semantic Segmentation with Mask-adapted CLIP\vspace{-0.5em}}

\author{Feng Liang\thanks{Work done during an internship at Meta Reality Labs.}~~\textsuperscript{\rm 1}, 
        Bichen Wu\textsuperscript{\rm 2}, 
        Xiaoliang Dai\textsuperscript{\rm 2},
        Kunpeng Li\textsuperscript{\rm 2}, 
        Yinan Zhao\textsuperscript{\rm 2}, 
        Hang Zhang\thanks{Work done while at Meta Reality Labs.}~~\textsuperscript{\rm 3}, \\
        Peizhao Zhang\textsuperscript{\rm 2}, 
        Peter Vajda\textsuperscript{\rm 2}, 
        Diana Marculescu\textsuperscript{\rm 1}\\
\textsuperscript{\rm 1}The University of Texas at Austin, \textsuperscript{\rm 2}Meta Reality Labs, \textsuperscript{\rm 3}Cruise \\
\texttt{\{jeffliang,dianam\}@utexas.edu}, \texttt{\{wbc,stzpz,vajdap\}@meta.com}\\
\texttt{\href{https://jeff-liangf.github.io/projects/ovseg}{https://jeff-liangf.github.io/projects/ovseg}}
}
\maketitle

\begin{abstract}
Open-vocabulary semantic segmentation aims to segment an image into semantic regions according to text descriptions, which may not have been seen during training. 
Recent two-stage methods first generate class-agnostic mask proposals and then leverage pre-trained vision-language models, \eg, CLIP, to classify masked regions. 
We identify the performance bottleneck of this paradigm to be the pre-trained CLIP model, since it does not perform well on masked images.
To address this, we propose to finetune CLIP on a collection of masked image regions and their corresponding text descriptions. We collect training data by mining an existing image-caption dataset (\eg, COCO Captions), using CLIP to match masked image regions to nouns in the image captions.
Compared with the more precise and manually annotated segmentation labels with fixed classes (\eg, COCO-Stuff), we find our noisy but diverse dataset can better retain CLIP's generalization ability. 
Along with finetuning the entire model, we utilize the ``blank'' areas in masked images using a method we dub \textit{mask prompt tuning}.
Experiments demonstrate mask prompt tuning brings significant improvement without modifying any weights of CLIP, and it can further improve a fully finetuned model. 
In particular, when trained on COCO and evaluated on ADE20K-150, our best model achieves 29.6\% mIoU, which is +8.5\% higher than the previous state-of-the-art.
For the first time, open-vocabulary \emph{generalist} models match the performance of supervised \emph{specialist} models in 2017 without dataset specific adaptations. 
\end{abstract}

\vspace{-0.5em}
\section{Introduction}
\label{sec:intro}
Semantic segmentation aims to group pixels into meaningful regions with corresponding semantic categories.
Although remarkable progress has been made~\cite{long2015fcn,chen2017deeplab,chen2018encoder,zhao2017pyramid,cheng2021maskformer}, modern semantic segmentation models are mainly trained with pre-defined categories, failing to generalize to unseen classes.
On the contrary, humans understand scenes in an open-vocabulary manner, typically with thousands of categories~\cite{biederman1987recognition}.
To approach human-level perception, this paper studies open-vocabulary semantic segmentation where the model segments an image by arbitrary categories described by texts.

\begin{figure*}[t]
    \centering
	\includegraphics[width=1.9\columnwidth]{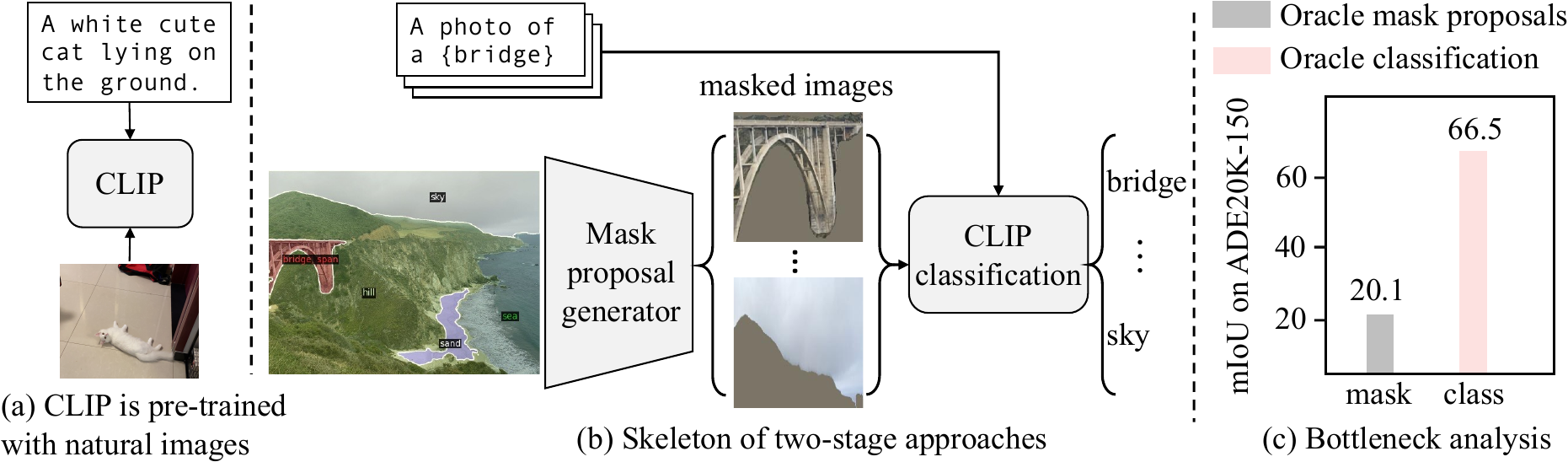}
	\caption{(a) CLIP is pre-trained with natural images with little data augmentation.
	(b) Two-stage open-vocabulary semantic segmentation approaches first generate class-agnostic mask proposals and then leverage pre-trained CLIP to do open-vocabulary classification.
	The input of the CLIP model is cropped masked images, which have huge domain gap from the natural images.
	(c) Our analysis reveals that pre-trained CLIP does not work well on masked images.
	}
	\vspace{-0.8em}
	\label{fig:main_figure}
\end{figure*}

Vision-language models, \eg, CLIP~\cite{radford2021clip}, learn rich multi-modal features from billion-scale image-text pairs.
Witnessing its superior open-vocabulary classification ability, prior works propose to use pre-trained vision-language models for open-vocabulary segmentation~\cite{li2022lseg,xu2021simple,ding2022zegformer,ghiasi2021openseg}.
Among them, two-stage approaches have shown great potential: they first generate class-agnostic mask proposals and then leverage pre-trained CLIP to perform open-vocabulary classification (see Figure~\ref{fig:main_figure}(b)).
Their success relies on two assumptions: (1) the model can generate class-agnostic mask proposals (2) pre-trained CLIP can transfer its classification performance to masked image proposals.

To examine these two assumptions, we conduct the following analysis. First, we assume an ``oracle'' mask generator and an ordinary CLIP classifier. We use ground-truth masks as region proposals, and feed masked images to a pre-trained CLIP for classification. This model only reaches an mIoU of 20.1\% on the ADE20K-150 dataset. Next, we assume an ``oracle'' classifier but an ordinary mask proposal generator -- a MaskFormer (\cite{cheng2021maskformer}) pre-trained on the COCO dataset. We first extract masked region proposals, then compare each region with ground-truth object masks, find the object with the highest overlap, and assign the object label to this extracted region. This model, despite imperfect region proposals, reaches a significantly higher mIoU of 66.5\%. 

This analysis clearly shows that pre-trained CLIP can \emph{not} perform satisfactory classification over masked images, and it is the performance bottleneck of two-stage open-vocabulary segmentation models. We hypothesize that this is caused by the significant domain gap between masked images and CLIP's training images. CLIP is pre-trained on natural images with minimal data augmentation \cite{radford2021clip}. On the other hand, mask proposals are cropped and re-sized from original images, and are further corrupted by noisy segmentation masks, see examples in Figure \ref{fig:main_figure} (b). 

To address this, we propose to adapt CLIP by finetuning it on masked images and corresponding text labels. One direct solution is to use segmentation labels, \eg, from the COCO-stuff dataset. However, this leads to bad generalization to unseen classes (Section \ref{sec:ablation_mask_category_pairs}). Such manually annotated masks are accurate but classes are limited to a closed set (\eg, 171 classes for COCO-stuff). We hypothesize that the lack of text diversity causes the finetuned CLIP to lose the generalization ability to open vocabulary concepts. Instead, we collect training data by mining an existing image-caption dataset (\eg, COCO Captions). Given an image-caption pair, we first extract nouns in the caption, and generate class-agnostic masked region proposals using a pre-trained segmentation model. Then, with a pre-trained CLIP model, we assign the best-matching proposal to each extracted noun. By learning from this weakly-supervised alignments between masked images and novel categories, the adapted CLIP better retains its generalization ability for open vocabulary classification.

The next question is how to effectively finetune CLIP? The most notable difference between a masked image and a natural image is that background pixels in a masked image are masked out, leading to many blank areas, which will be converted to ``zero tokens'' when feeding to CLIP transformers. Such tokens not only contain no useful information, but also bring domain distribution shift to the model (since such tokens don't exist in natural images) and cause performance degradation. To mitigate this, we propose mask prompt tuning, \'a la visual prompt tuning \cite{jia2022vpt}. When tokenizing a masked image, we replace the ``zero tokens'' with learnable prompt tokens. During finetuning, we either train prompts only and freeze CLIP's weights, or train both of them. We find that mask prompt tuning alone significantly improves CLIP's performance on masked images. This is a crucial property for multi-task scenarios where we cannot change CLIP's weight since it is shared with other tasks. When combined with full model finetuning, mask prompt tuning can further improve the performance by a non-trivial margin (see Section~\ref{sec:mask_prompt_learning}).

In our experiments, we measure the open-vocabulary segmentation performances on holdout segmentation datasets in a ``zero-shot'' manner -- we do not adapt the model for each evaluation dataset. We train our model using COCO-stuff~\cite{caesar2018cocostuff} dataset with captions from ~\cite{chen2015cococaption}, and test on challenging ADE20K (A-150, A-847 for 150/846 categories)~\cite{zhou2019ade}, Pascal Context (PC-59, PC-459 for 59/459 categories)~\cite{mottaghi2014pascalcontext} and PASCAL VOC (PAS-20)~\cite{everingham2010pascalvoc}.
Our best model achieves 29.6\% mIoU on A-150, which is +8.5\% than the state-of-the-art OpenSeg~\cite{ghiasi2021openseg} under the same setting.
On more challenging A-847 and PC-459, our model sets up a new state-of-the-art of 9.0\%, 12.4\% mIoU, surpassing the previous best solution by +2.7\% and 3.4\%.
Moreover, for the first time, we show open-vocabulary \emph{generalist} models can match the performance of supervised \emph{specialist} models~\cite{long2015fcn,chen2017deeplab,zoph2020rethinking} in 2017 without dataset specific adaptations.

In summary our contributions include: (1) Our analysis reveals the pre-trained CLIP does \emph{not} perform well on mask proposals, making it the performance bottleneck of two-stage approaches.
(2) We collect diverse mask-category pairs from captions to adapt CLIP for masked images and retain its generalization ability.
(3) We propose mask prompt tuning specifically for masked image adaptation. This method does not change CLIP's weight, enabling multi-task weight sharing.
(4) For the first time, we show open-vocabulary \emph{generalist} models can match the performance of supervised \emph{specialist} models in 2017 without dataset specific adaptations.

\section{Related Work}

\begin{figure*}[t]
    \centering
	\includegraphics[width=1.8\columnwidth]{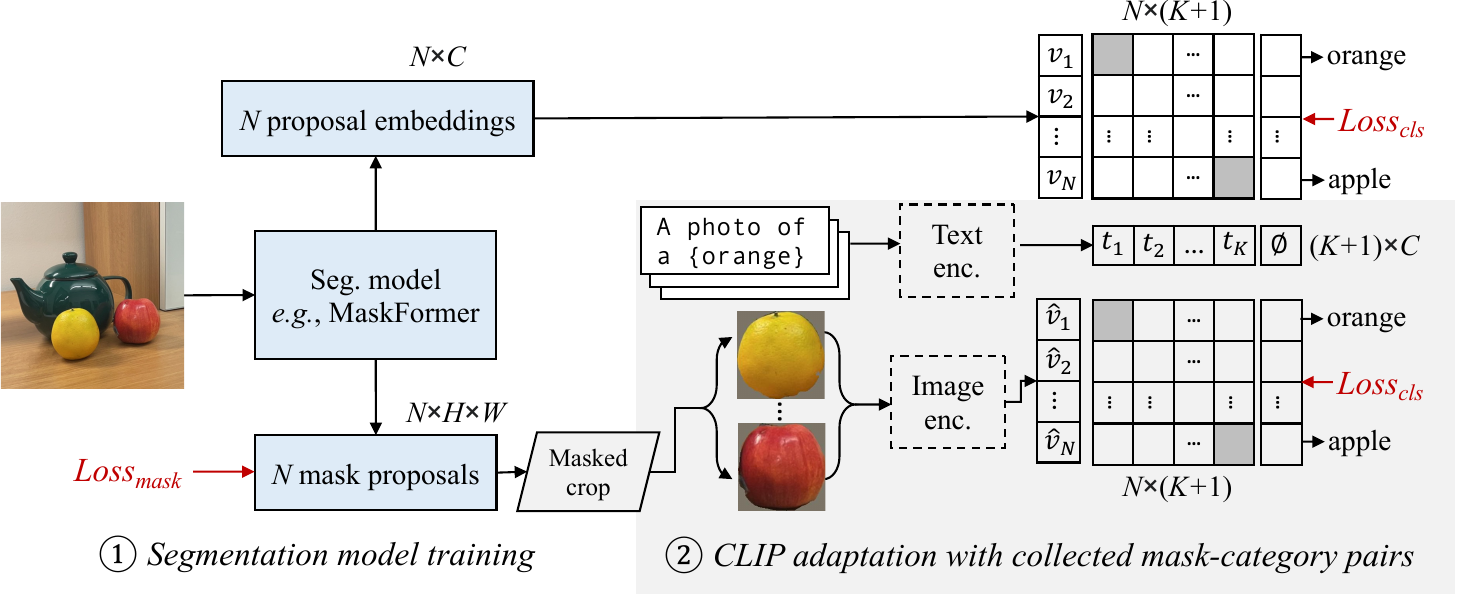}   
    \vspace{-1em}
	\caption{Two-stage approaches consist of one segmentation model, \eg, MaskFormer, and one CLIP model. 
	Firstly, the modified MaskFormer is trained with CLIP’s text embeddings so as to perform open-vocabulary segmentation. (Section~\ref{sec:two_stage_approach}).
	We then use the pre-trained segmentation model to generate class-agnostic proposals and align proposals with extracted nouns from corresponding captions (Section~\ref{sec:collecting_diverse_pairs}). After collecting diverse mask-category pairs, we finetune CLIP with the proposed mask prompt tuning (Section~\ref{sec:method_mask_prompt_learning}).
	}
	\vspace{-1em}
	\label{fig:two-stage-method}
\end{figure*}

\textbf{Pre-trained vision-language models}~\cite{radford2021clip,jia2021align,li2021declip,wu2021otter} connect the visual concepts with textual description.
Pre-trained CLIP~\cite{radford2021clip} has strong open-vocabulary classification ability, \ie, classifying an image with arbitrary categories described by language.
Pre-trained CLIP has empowered many computer vision tasks with the language ability, such as image manipulation~\cite{patashnik2021styleclip}, image generation~\cite{crowson2022vqganclip}, object detection~\cite{gu2021vild,zhong2022regionclip} and image segmentation~\cite{li2022lseg,xu2021simple,ding2022zegformer,ghiasi2021openseg, ding2022maskclip,luo2022segclip,kim2023zegot,xu2023side}.
Our work is similar to RegionCLIP~\cite{zhong2022regionclip}, which adapts CLIP for object detection by finetuning on region proposals. Our method differs from RegionCLIP since (1) we adapt CLIP to process masked images while RegionCLIP processes complete region crops; (2) We leverage blank areas in masked images and propose mask prompt tuning, which adapts CLIP without changing its weights. This enables sharing CLIP's weight with other tasks in multi-task scenarios. This is not supported by RegionCLIP.

\textbf{Open-vocabulary segmentation} aims to understand an image with arbitrary categories described by texts. Pioneering work ZS3Net~\cite{bucher2019zs3net} uses generative models to synthesize pixel-level features by word embeddings of unseen class. SPNet~\cite{xian2019spnet} utilizes the word embeddings, \eg, word2vec~\cite{mikolov2013word2vec}, to align the semantic meaning with visual features.
GroupViT~\cite{xu2022groupvit} groups segmentation masks directly from text supervision. 
More recently, researchers propose to leverage the pre-trained CLIP~\cite{radford2021clip} for open-vocabulary semantic segmentation.
LSeg~\cite{li2022lseg} aligns pixel embeddings to the text embedding of the corresponding semantic class, which is generated by CLIP's text encoder. 
Unlike pixel-level LSeg, OpenSeg~\cite{ghiasi2021openseg} proposes to align the segment-level visual features with text embedding via region-word grounding.
Our approach falls into the category of two-stage approaches, such as ZSSeg~\cite{xu2021simple} and ZegFormer~\cite{ding2022zegformer}: they first generate class-agnostic mask proposals and then utilize pre-trained CLIP to perform open-vocabulary classification.
Unlike ZSSeg and ZegFormer which directly use the original CLIP for masked image classification, we adapt CLIP to improve performance.

\textbf{Prompt tuning} is a strategy to adapt large-scale pre-trained models to new tasks. The idea originated from natural language processing~\cite{liu2021prompt,li2021prefix,lester2021power}, and recent work extends prompt tuning to computer vision.
CoOp~\cite{zhou2022coop} pre-appends the category words with learnable vectors to adapt CLIP for many recognition tasks.
The textual prompt tuning is also widely used in open-vocabulary object detection~\cite{du2022learning} and semantic segmentation~\cite{xu2021simple}.
Our mask prompt tuning is more relevant to prompt tuning in the visual domain~\cite{bahng2022exploring,jia2022vpt} where learnable vectors are applied to the image domain. Unlike visual prompt tuning~\cite{jia2022vpt} that inserts additional tokens before the actual image tokens, we \textit{replace} masked tokens with learnable prompts. 
Furthermore, mask prompt tuning brings additional improvement over a fully finetuned model (Section \ref{sec:mask_prompt_learning}). Such additional improvements have not been reported by prior work.

\section{Method}
\label{sec:method}

In this section, we first revisit the two-stage open-vocabulary segmentation methods~\cite{xu2021simple,ding2022zegformer}.
Then we discuss how to obtain a dataset of mask-category pairs to finetune CLIP. Last, we discuss the proposed mask prompt tuning technique to adapt CLIP for masked images.

\subsection{Two-stage models for open-vocabulary semantic segmentation}
\label{sec:two_stage_approach}
Our two-stage open-vocabulary semantic segmentation model is shown in Figure~\ref{fig:two-stage-method}. It consists of a segmentation model that generates mask proposals, and an open vocabulary classification model. 

Following ~\cite{xu2021simple,ding2022zegformer}, we choose MaskFormer~\cite{cheng2021maskformer} as the segmentation model. Unlike per-pixel segmentation models~\cite{long2015fcn,chen2017deeplab}, MaskFormer predicts a set of $N$ mask proposals and corresponding class predictions. Each proposal is represented by an $H \times W$ binary mask, indicating the location of the target object. The class prediction is a $C$-dimensional distribution, where $C$ is the number of classes in the training set. Following \cite{xu2021simple}, we modify MaskFormer such that for each mask, it generates a $C$-dimensional proposal embedding, where $C$ is the embedding dimension of a CLIP model (512 for ViT-B/16 and 768 for ViT-L/14). This change allows MaskFormer to perform open-vocabulary segmentation. Specifically, suppose we would like to classify the mask to $K$ categories, we can first use a CLIP model's text encoder to generate K text embeddings for each class as $\{t_k | t_k \in \mathbf{R}^C\}_{k=1,\cdots, K}$. Next, we compare each mask embedding $v_i$ with the text embedding, and predict the class-$k$ probability as $p_{i, k} = \exp(\sigma(v_i, t_k)/\tau) / \sum_k (\exp(\sigma(v_i, t_k)/\tau))$. 
Here $\sigma(\cdot, \cdot)$ denotes the cosine similarity between two embedding vectors, and $\tau$ is the temperature coefficient \cite{radford2021clip}. We train the modified MaskFormer on the COCO-Stuff dataset \cite{caesar2018cocostuff} with 171 classes. We use CLIP's text encoder to process class names to generate text embeddings. We also append a learnable embedding $\emptyset$ to represent the category of ``no object". For other training settings, we follow the original MaskFormer \cite{cheng2021maskformer}. 

Note that the mask proposal generator trained this way is not strictly ``class-agnostic", as the definition of an object is determined by the class definitions in the training set. For example, if the training set only contains "person" as a class, it is not likely the model will automatically segment a person into ``face", ``hand", ``body", or finer body parts. How to train a general and class agnostic model to generate mask proposals is an important topic but is beyond the scope of this paper.

In addition to MaskFormer's prediction, following \cite{ding2022zegformer,xu2021simple}, we add a parallel prediction branch using CLIP. MaskFormer generates mask proposals $\{M_i | M_i \in \{0, 1\}^{H\times W}\}_{i=1,\cdots, N}$ where $1$ and $0$ denotes foreground and background. For each mask, we select a tight bounding box that includes all foreground pixels, crop the image, mask out backgrounds, and re-size to CLIP's resolution. We feed mask proposal-$i$ to CLIP and compute class-$k$ probability as $\hat{p}_{i,k}$. We ensemble both predictions to compute final prediction as $p_{i, k}^{(1-\lambda)} * \hat{p}_{i, k}^{\lambda}$
where $\lambda \in [0, 1]$. We fuse mask-wise predictions to semantic segmentation using MaskFormer's fusion module. 

As discussed in Section~\ref{sec:intro} and Figure \ref{fig:main_figure} (c), our analysis show that CLIP does \emph{not} work well on such masked images. Specifically, CLIP is trained on natural images with little data augmentation \cite{radford2021clip}. However, masked images as shown in Figure \ref{fig:main_figure} (b) contain a lot of  ``blank regions". Such a significant domain gap makes it difficult for CLIP to transfer its classification performance. We also tried cropping the proposals without masking out background pixels. However, we observe worse performance (see Appendix).
We conjecture that keeping background pixels makes it more confusing for CLIP to correctly classify the foreground.

\subsection{Collecting diverse mask-category pairs from captions}
\label{sec:collecting_diverse_pairs}

To adapt CLIP to better process masked images, we propose to finetune CLIP on a dataset consisting of masked image and text pairs. One direct solution is to leverage manually annotated segmentation labels, \eg, from COCO-Stuff. Such labels are accurate but have a closed set of categories. We try this solution and collect 965K mask-category pairs spanning 171 classes (\eg, banana, orange) from COCO-Stuff. 
Then we finetune the CLIP's image encoder, while freezing the text encoder, following ~\cite{zhong2022regionclip}.
However, we observe that this naive approach limits the generalization ability of CLIP, as the performance drops if there are more unseen classes (see Section~\ref{sec:ablation_mask_category_pairs}). We hypothesize that due to the limited text vocabulary, the finetuned CLIP over-fits to the 171 classes, losing the ability to generalize to unseen categories. 

\begin{figure}[t]
    \begin{center}
    \includegraphics[width=\linewidth]{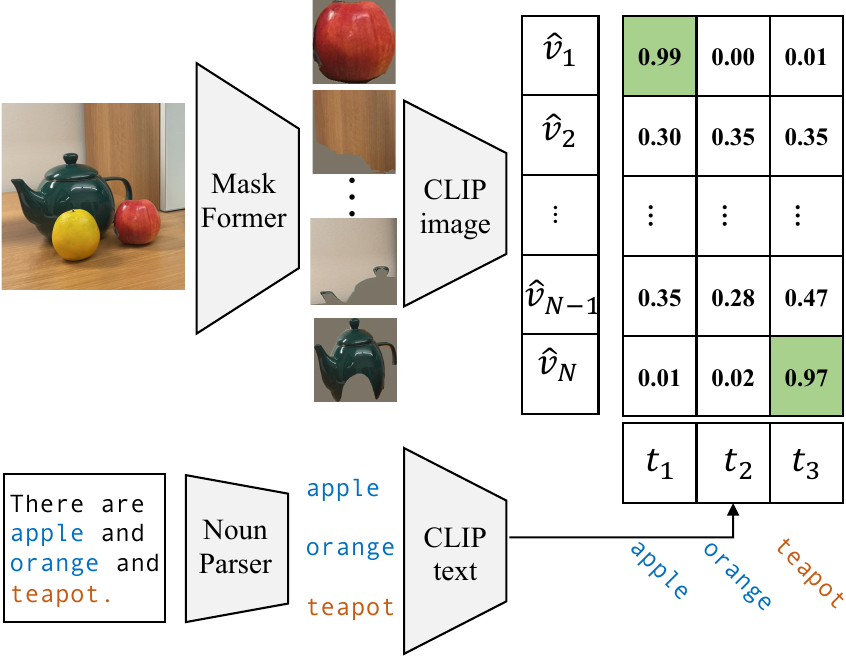}
    \end{center}
    \vspace{-1em}
        \caption{For the given image-cation pair,
        only \texttt{"apple"} and \texttt{"orange"} are categories in COCO.
        By extracting nouns from captions, we can also get a novel \texttt{"teapot"} category.}
        \vspace{-1em}
    \label{fig:diverse_pairs}
\end{figure}

\begin{figure*}[t]
    \centering
	\includegraphics[width=1.8\columnwidth]{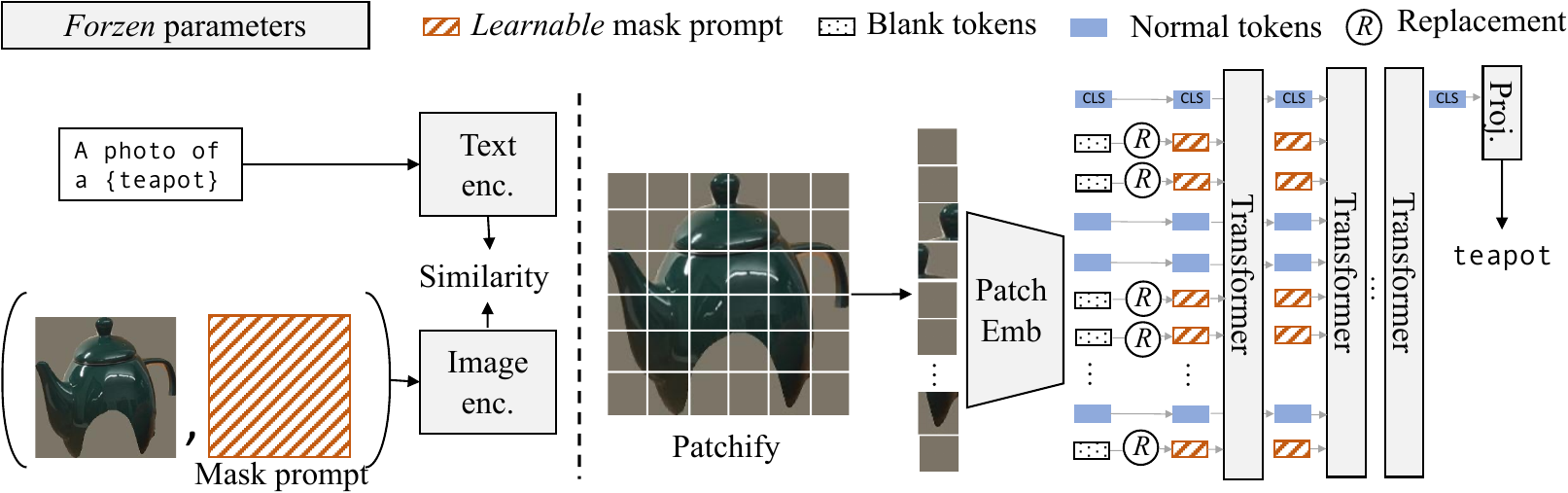}   
    \vspace{-0.2cm}
	\caption{The proposed mask prompt tuning can adapt CLIP to masked images without changing its weights. We replace the zero tokens from masked patches with learnable mask prompts.}
	\label{fig:mask_prompt_tuning}
	\vspace{-0.3cm}
\end{figure*}

Compared with segmentation labels, image captions contain much richer information about images and involve a much larger vocabulary. For example, in Figure~\ref{fig:diverse_pairs}, the image caption is \texttt{"There are apple and orange and teapot."}. Though \texttt{"apple"} and \texttt{"orange"} are valid classes in COCO-Stuff, other concepts are not valid classes and are ignored. 

Based on this observation, we designed a self-labeling strategy~\cite{ghiasi2021openseg,zhong2022regionclip} to extract mask-category pairs. As in Figure \ref{fig:diverse_pairs}, given an image, we first use a pre-trained MaskFormer to extract masked proposals. Meanwhile, from the corresponding image caption, we extract all nouns using an off-the-shelf language parser~\cite{bird2009nltk}, and treat them as potential classes. Then, we use CLIP to pair the most matching mask proposal to each class. From COCO-Captions ~\cite{chen2015cococaption}, we collect 1.3M mask-category pairs with 27K unique nouns using 5 captions per image, or 440K pairs with 12K nouns using 1 caption per image. Experiments show this noisy but diverse mask-category dataset leads to significantly better performance than manual segmentation labels (see Section~\ref{sec:ablation_mask_category_pairs}).

\subsection{Mask prompt tuning}
\label{sec:method_mask_prompt_learning}
After collecting the dataset, a natural question is how to finetune CLIP effectively? The most notable difference between a masked image and a natural image is that background pixels in a masked images are set to zeros, leading to many ``blank areas''. When feeding masked images to CLIP, images will be divided into non-overlapping patches and subsequently tokenized. Those blank areas will then become zero tokens. Such tokens not only contain no useful information but also bring domain distribution shift to the model (since such tokens don't exist in natural images) and cause performance degradation. To mitigate this, we propose a technique called \textit{mask prompt tuning}, \`a la visual prompt tuning~\cite{jia2022vpt}. Specifically, when feeding into CLIP, a masked image will be tokenized to a tensor $T \in \mathbf{R}^{N_p \times E}$, where $N_p$ is the number of patches, and $E$ is the token dimension. The masked image also comes with a condensed binary mask $M_p \in \{0, 1\}^{N_p}$, where each element indicates whether a given patch is kept or masked out. 
Only when all the pixels within the patch are entirely masked, is the patch treated as a masked token. The intuition is that the boundary pixels, which usually exist in partially masked patches, are crucial for region classification.
We allocate a learnable tensor representing prompt tokens as $P \in \mathbf{R}^{N_p \times E}$. Finally, the final input to the transformer is computed as $T \otimes M_p + P \otimes (1 - M_p) $, where $\otimes$ denotes element-wise multiplication. Following the ``deep prompts" in ~\cite{jia2022vpt}, we can add such prompt tokens to deeper layers of the transformer. This is also illustrated in Figure \ref{fig:mask_prompt_tuning}. 

Compared with fully finetuning the entire model \cite{zhong2022regionclip}, mask prompt tuning has several advantages. First, it is specifically designed for segmentation tasks, where parts of input images are masked. Next, compared with full model finetuning, the amount of trainable parameters in mask prompt tuning is orders of magnitude smaller, leading to much better training efficiency. Moreover, as a foundational model, CLIP may be simultaneously used for many tasks, and we may not be allowed to tune CLIP's weights. Mask prompt tuning does not require changing weights of CLIP, thus is suitable for such multi-task scenarios. Lastly, our experiments show that mask prompt tuning alone leads to significant improvement.
And if applied together with full model finetuning, it can further improve the open-vocabulary segmentation performance (Section~\ref{sec:mask_prompt_learning}).

\vspace{-0.5em}
\section{Experiments}

\begingroup
\setlength{\tabcolsep}{4pt} 
\begin{table*}[t]
\vspace{-1em}
\caption{The mIoU results of open-vocabulary generalist models and supervised specialist models. 
Results for SPNet and ZS3Net on PAS-20 are reported from~\cite{li2022lseg}.
Results for ZegFormer on PAS-20 are recalculated by us.
SimBaseline~\cite{xu2021simple}, ZegFormer~\cite{ding2022zegformer} and OpenSeg~\cite{ghiasi2021openseg} are using the same COCO images, \ie, the 2017 splits with 118K images, but with different annotations. COCO-Stuff-156/171 denotes using COCO Stuff mask annotations of 156/171 categories.
Under the R101c model scale, our model significantly outperforms other open-vocabulary models.
Our largest Swin-Base model can match the performance of some supervised specialist models in 2017.
}
\vspace{-0.1cm}
\small
\centering
\begin{tabular}{l|ll|rrrrr}
\toprule  
  method  & backbone & training dataset  & A-847 & PC-459 & A-150 & PC-59 & PAS-20 \\
  \midrule \midrule
  \multicolumn{8}{c}{\textit{Open-vocabulary generalist models}} \\
  SPNet~\cite{xian2019spnet} & R-101 & PASCAL-15 & - & -& - & 24.3 & 18.3 \\
  ZS3Net~\cite{bucher2019zs3net} & R-101 & PASCAL-15 & - & -& - & 19.4 & 38.3 \\
  LSeg~\cite{li2022lseg} & R-101 & PASCAL-15 & - & -& - & - & 47.4 \\
  LSeg+~\cite{ghiasi2021openseg} & R-101 & COCO Panoptic & 2.5 & 5.2 & 13.0 & 36.0 & 59.0 \\
  SimBaseline~\cite{xu2021simple} & R-101c & COCO-Stuff-156 & - & - & 15.3 & - & 74.5\\
  ZegFormer~\cite{ding2022zegformer} & R-50 & COCO-Stuff-156 & - & - & 16.4 & - & 80.7 \\
  OpenSeg~\cite{ghiasi2021openseg} & R-101 & COCO Panoptic & 4.0 & 6.5 & 15.3 & 36.9 & 60.0\\
  \ours  (Ours) & R-101c & COCO-Stuff-156  & 7.0 & 10.4 & 24.0 & 51.7 & 89.2 \\
  \ours  (Ours) & R-101c & COCO-Stuff-171  & \textbf{7.1} & \textbf{11.0} & \textbf{24.8} & \textbf{53.3} & \textbf{92.6} \\
  \midrule
  LSeg+~\cite{ghiasi2021openseg} & Eff-B7 & COCO Panoptic & 3.8  &   7.8   &  18.0   &  46.5  & - \\
  OpenSeg~\cite{ghiasi2021openseg} & Eff-B7 & COCO Panoptic & 6.3 & 9.0 & 21.1 & 42.1  & - \\
  \ours  (Ours) & Swin-B & COCO-Stuff-171  & \textbf{9.0} & \textbf{12.4}  & \textbf{29.6} & \textbf{55.7} & \textbf{94.5} \\
  \midrule
  \multicolumn{8}{c}{\textit{\textcolor{gray}{Supervised specialist models}}} \\
  \textcolor{gray}{FCN~\cite{long2015fcn}} & \textcolor{gray}{FCN-8s} & \textcolor{gray}{Same as test} & - & - & \textcolor{gray}{29.4} & \textcolor{gray}{37.8}  & - \\
  \textcolor{gray}{Deeplab~\cite{chen2017deeplab}} & \textcolor{gray}{R-101} & \textcolor{gray}{Same as test} & - & - & - & \textcolor{gray}{45.7}  & \textcolor{gray}{77.7}  \\
  \textcolor{gray}{SelfTrain~\cite{zoph2020rethinking}} & \textcolor{gray}{Eff-L2} & \textcolor{gray}{Same as test} & - & - & - & -  & \textcolor{gray}{90.0}  \\
  \textcolor{gray}{MaskFormer~\cite{cheng2021maskformer}} & \textcolor{gray}{R-101c} & \textcolor{gray}{Same as test} & \textcolor{gray}{17.4} & - & \textcolor{gray}{46.0} & -  & - \\
  \bottomrule
\end{tabular}
\vspace{-0.2cm}
\label{tab:sota}
\end{table*}
\endgroup

\subsection{Experimental setup}
\label{sec:setup}
\textbf{Training Dataset} We train our model on the COCO dataset~\cite{lin2014coco}. We first train the modified MaskFormer using the segmentation labels from COCO-Stuff~\cite{caesar2018cocostuff}. Next, we finetune CLIP on the mask-category dataset that we obtained from COCO Captions~\cite{chen2015cococaption}.
There are 118k training images labeled with 171 valid categories in the dataset, ranging from things (\eg, orange, car) to stuffs (\eg, sky, road). 
If not specified otherwise, we use all the 171 categories data during training.

\textbf{Evaluation Dataset} Our open-vocabulary model is able to perform zero-shot segmentation on arbitrary datasets without dataset-specific adaption. Thus, we test our model on challenging ADE20K~\cite{zhou2019ade}, Pascal VOC~\cite{everingham2010pascalvoc} and Pascal Context~\cite{mottaghi2014pascalcontext} datasets.
ADE20K is a densely pixel-wise annotated dataset for scene understanding, which spans diverse annotations of indoor and outdoor scenes. 
There are 2K images in its validation set.
We choose two versions of categories, one with 150 frequently used categories (A-150) and one with more diverse 847 categories (A-847). 
Pascal VOC is a classical dataset for segmentation. We evaluate on the 1.5K validation images with 20 categories (PAS-20).
Pascal Context is a set of additional annotations for PASCAL VOC 2010. 
It goes beyond the original PASCAL semantic segmentation task by providing annotations for the whole scene. 
There are 5K images in its validation set.
We also choose two versions of categories, one with 59 frequently used categories (PC-59) and one with the whole 459 categories (PC-459).

\begin{table*}[t]
\vspace{-1em}
\caption{Ablation on mask-category pairs.
The baseline is MaskFormer Swin-Base with original CLIP ViT-L/14.
The masks come from ground-truth (GT) or generated proposals.
The category nouns come from ground-truth (GT) classes or captions.
We also calculate the statistics (number of pairs and unique nouns) of collected pairs.
}
\vspace{-0.2cm}
\small
\centering
\begin{tabular}{c|cc|cc|ccc}
\toprule
\multirow{2}{*}{Case} & \multicolumn{2}{c|}{Source} & \multicolumn{2}{c|}{Statistics} & \multirow{2}{*}{A-847} & \multirow{2}{*}{A-150} & \multirow{2}{*}{PC-59} \\
\cmidrule(lr){2-3} \cmidrule(lr){4-5}
                      & Mask        & Category     & Pairs          & Unique nouns         &                        &                        &                        \\
\midrule \midrule
Baseline              & -           & -            & -              & -             &    7.3                    &   21.8                     &    51.4                    \\
(1)                   & GT          & GT           & 965K           & 171           &   5.3 \color{red}{(-2.0)}              &  23.0 \color{ForestGreen}{(+1.2)}           &     \textbf{57.3} \color{ForestGreen}{(+5.9)}               \\
(2)                   & GT          & 1 caption    & 440K           & 12K           &   7.9 \color{ForestGreen}{(+0.6)}             &   24.2 \color{ForestGreen}{(+2.4)}                       &   53.2  \color{ForestGreen}{(+1.8)}               \\
(3)                   & proposals   & 1 caption    & 440K           & 12K           &   \textbf{8.8} \color{ForestGreen}{(+1.5)}                 &  \textbf{28.8} \color{ForestGreen}{(+7.0)}                      &         55.7  \color{ForestGreen}{(+4.3)}              \\
(4)                   & proposals   & 5 captions   & 1.3M           & 27K           &   \textbf{8.8} \color{ForestGreen}{(+1.5)}                    &  28.6  \color{ForestGreen}{(+6.8)}                &     55.5  \color{ForestGreen}{(+4.1)}           \\
\bottomrule
\end{tabular}
\vspace{-1em}
\label{tab:mask_category_pairs}
\end{table*}

\textbf{Implementation Details} As indicated before, our model consists of two part: one segmentation model based on MaskFormer~\cite{cheng2021maskformer} and one mask-adapted CLIP model~\cite{radford2021clip}.
The final class prediction is ensemble of MaskFormer's prediction and CLIP's prediction. The ensemble weight $\lambda$ can be found in Appendix.
For the segmentation model, we have two backbone choices, ResNet-101c~\cite{chen2017deeplab} and Swin-Base~\cite{liu2021swin}. For the CLIP model, we have two choices: ViT-B/16 and ViT-L/14~\cite{dosovitskiy2020vit}.
We detail our largest model setting here, while the training recipe of the R101c model can be found in Appendix.
For Swin-Base segmentation model, the backbone weights are initialized from an ImageNet-21K pre-trained model.
We use AdamW~\cite{loshchilov2017adamw} optimizer with the poly learning rate schedule \cite{chen2017deeplab}.
The initial learning rate and weight decay are set to $6 \cdot 10^{-5}$ and $10^{-2}$, respectively.
We use a crop size of $640 \times 640$, a batch size of 32 and train the model for 120K iterations.
For data augmentations and other hyper-parameters, we mainly follow the setting of ~\cite{cheng2021maskformer}.

For adapting CLIP ViT-L/14 model, we utilize the OpenCLIP~\cite{ilharco_gabriel_2021_5143773} implementation.
After collecting 440K mask-category pairs from captions (see Section~\ref{sec:collecting_diverse_pairs}), we propose three ways to adapt CLIP: mask prompt tuning (MPT) only, full model fine-tuning (FT) only and joint MPT + FT.
For MPT only, we initialize the CLIP model with official OpenAI weights~\cite{radford2021clip} and the learnable tokens are randomly initialized. 
We also use the deep prompts as proposed in ~\cite{jia2022vpt}.
The prompt depth is set to 3 if not specified otherwise.
The training optimizer is AdamW with initial learning rate $2 \cdot 10^{-2}$ and weight decay $0$.
The cosine annealing scheduler is adopted to adjust the learning rate.
The model is trained with input size of $224 \times 224$, a batch size of 256 for 5 epochs.
For FT only, we keep similar training procedure but with a much lower learning rate $5 \cdot 10^{-6}$ and larger weight decay $0.2$.
For MPT + FT, we first initialize the CLIP with fully finetuned model and then apply the mask prompt tuning over it, which we fined more stable and effective (see Appendix)
All other hyper-parameters are the same with MPT only.  
The text encoder of CLIP is frozen in all our experiments.

\subsection{Main results on open vocabulary semantic segmentation}

\textbf{\ours achieves best performance among open-vocabulary models.}
We conduct the comparison with other open-vocabulary generalist models using the common ResNet-101 (R-101) model scale in Table~\ref{tab:sota}.
We use R-101c~\cite{chen2017deeplab}, which replaces the first $7\times7$ convolution layer of R-101 with 3 consecutive $3\times3$ convolutions and which is popular in the semantic segmentation community.
If not specified otherwise, our best performance is achieved using joint mask prompt tuning and fine-tuning (see Section~\ref{sec:mask_prompt_learning}).
First of all, compared with per-pixel approaches (SPNet~\cite{xian2019spnet}, ZS3Net~\cite{bucher2019zs3net}, LSeg~\cite{li2022lseg} and LSeg+~\cite{ghiasi2021openseg}), proposal-based approaches (OpenSeg~\cite{ghiasi2021openseg}, SimBaseline~\cite{xu2021simple} and ZegFormer~\cite{ding2022zegformer}) show better performance.
Our \ours also falls into the proposal-based category.
Compared with other proposal-based approaches, our model shows significant improvements across all five benchmarks. 
In particular, our R101c model achieves 7.1\% and 11.0\% mIoU on challenging A-847 and PC-459, which even performs better than the EfficientNet-B7 based OpenSeg model.
We notice open-vocabulary segmentation is a new research problem, thus different approaches may use different experimental settings, such as different COCO annotations. 
Our experiments show different annotations result in relatively small performance differences: we only observe a 0.8\% mIoU drop on A-150 when changing COCO-Stuff-171 to COCO-Stuff-156.

\textbf{Largest \ours model sets up new SOTA results on zero-shot benchmarks.} 
When we scale up the model, our method can further achieve better results.
With Swin-Base (Swin-B) backbone and CLIP ViT-L/14, our model can achieve 29.6\% and 55.5\% mIoU on A-150 and Pascal PC-59, which is +8.5\% and +13.6\% higher than the SOTA zero-shot results. 
On the challenging A-847 and PC-459, our model sets up a new zero-shot state-of-the-art 9.0\% and 12.4\% mIoU.
We further detail the class-wise IoU of A-150 categories in Appendix.

\textbf{Open-vocabulary generalist models can match supervised specialist models in 2017.} 
We show our generalist model can achieve competitive performance without the need of any dataset specific training.
On the challenging A-150, our model achieves similar performance with fully supervised FCN-8s~\cite{long2015fcn}.
On the PAS-20, our model achieves 94.5\% mIoU, which is even +4.5\% than the SOTA specialist model~\cite{zoph2020rethinking}.
We note OVSeg is not directly comparable with supervised models because OVSeg is not trained on evaluation datasets. 
OVSeg also has different backbones and segmentation model architectures.
Thus, comparison with supervised models is for reference purposes only.
Our generalist model still underperforms the advanced specialist models, such as supervised MaskFormer~\cite{cheng2021maskformer}.

\subsection{Ablation study}

\begin{table}[t]
\caption{Ablation on mask prompt tuning (MPT) and full model tuning. 
The baseline is MaskFormer Swin-Base with CLIP ViT-L/14.
We report the zero-shot mIoU on representative ADE-847, ADE-150 and PC-59 datasets.
All the improvements are measured upon the baseline model.
}
\vspace{-0.2cm}
\small
\centering
\addtolength{\tabcolsep}{-1.5pt}
\begin{tabular}{c|cc|ccc}
\toprule
\multirow{2}{*}{case} & \multicolumn{2}{c|}{FT method}& \multirow{2}{*}{A-847} & \multirow{2}{*}{A-150} & \multirow{2}{*}{PC-59} \\
\cmidrule(lr){2-3}
                      & MPT              & full                                 &                          &                          &                        \\
\midrule \midrule
Baseline &     &  & 7.3     & 21.8    & 51.4  \\
(a)      & \checkmark   &   & 8.4 \color{ForestGreen}{(+1.1)}    & 26.5 \color{ForestGreen}{(+4.7)}   & 55.4 \color{ForestGreen}{(+4.0)}  \\
(b)      &     & \checkmark  &  8.8 \color{ForestGreen}{(+1.5)}    & 28.8 \color{ForestGreen}{(+7.0)}   & \textbf{55.7} \color{ForestGreen}{(+4.3)} \\
(c)      & \checkmark   & \checkmark  & \textbf{9.0} \color{ForestGreen}{(+1.7)}    & \textbf{29.6} \color{ForestGreen}{(+7.8)}   & \textbf{55.7} \color{ForestGreen}{(+4.3)} \\
\bottomrule
\end{tabular}
\addtolength{\tabcolsep}{-1.5pt}    
\vspace{-1em}
\label{tab:mask_prompt_learning}
\end{table}

\begin{figure*}[t]
    \vspace{-0.5em}
    \centering
	\includegraphics[width=1.9\columnwidth]{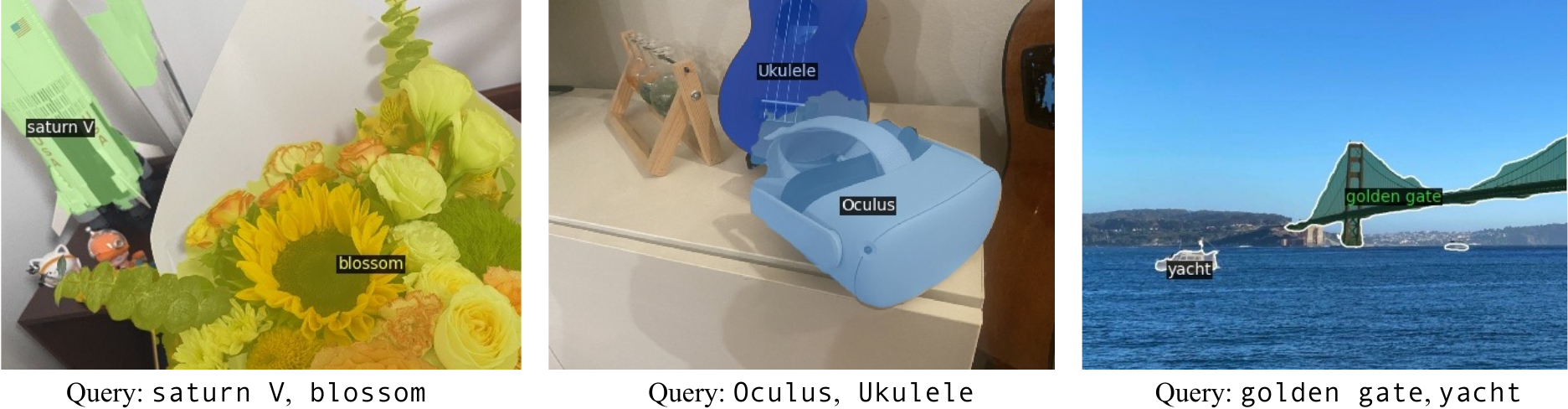}   
    \vskip -0.5em
	\caption{Open-vocabulary segmentation with user-defined queries. Our model accurately segments unseen categories, such as the \texttt{Saturn V} rocket, \texttt{Oculus} headset, and \texttt{Golden gate} bridge.}
	\label{fig:visualization}
	\vspace{-1em}
\end{figure*}

\subsubsection{Collecting mask-category pairs}
\label{sec:ablation_mask_category_pairs}
We discuss the impact of finetuning data in Table~\ref{tab:mask_category_pairs}.
The baseline model is MaskFormer Swin-Base with the original CLIP ViT-L/14.
Our initial trial (case (1)) is collecting ground-truth (GT) masks with supervised GT classes.
We can collecting 965K mask-category pairs with 171 unique nouns (the number of classes defined in COCO-stuff).
Then we finetune the CLIP model with the collected pairs.
We observe a -2.0\% performance drop on the A-847 dataset.
This is because the adapted CLIP is over-fitting to the 171 GT classes.
Although the model achieves good results on PC-59 (whose categories are highly overlapped with COCO-Stuff), it perform badly for more diverse concepts in A-847.
As detailed in Section~\ref{sec:collecting_diverse_pairs}, we propose to utilize captions~\cite{chen2015cococaption} to collect diverse mask-category pairs. 
After parsing the nouns in the caption, we match the nouns with GT masks (case (2)) or proposals (case (3)) generated by the baseline model.
By replacing the GT masks with proposals, the A-150 mIoU is significantly improved (from 24.2\% to 28.8\%)
We conjecture that many regions are not labeled as GT masks (see examples in Figure~\ref{fig:diverse_pairs}), and are therefore ignored. In contrast, the generated proposals (usually 100) can cover most of regions-of-interest in the image, leading to better performance.
If all the 5 captions per image are used (case (4)),
we observe a mild -0.2\% degradation on A-150 and PC-59
We hypothesis that 12K nouns are adequate for the CLIP to retain its open-vocabulary ability.
Thus, we choose to use 1
caption for efficiency purposes as it’s 5x faster in training
then using 5 captions.

\subsubsection{Mask prompt tuning}
\label{sec:mask_prompt_learning}
We ablate the effect of mask prompt tuning in Table~\ref{tab:mask_prompt_learning}. 
The baseline model is MaskFormer Swin-Base with CLIP ViT-L/14.
If we only use mask prompt tuning (case (a)), our model outperforms the baseline by a large +4.7\% and +4.0\% mIoU improvement on ADE-150 and PC-59, respectively.
Case (b) shows the result of full model fine-tuning.
Although it achieves the best accuracy, the trainable parameters are orders of magnitude higher. In contrast, the proposed mask prompt tuning only modifies the input without changing CLIP's weight.
Furthermore, mask prompt tuning can further improve over a fully finetuned model, as shown in case (c). 
Case (c) achieves 29.6\% mIoU ADE-150, which further improves the fully finetuned model by a considerable margin of +0.8\%.

\subsection{Discussions}

\subsubsection{Segmentation with user-defined queries.}

Our method allows users to define arbitrary queries and search the query in the image, see Figure~\ref{fig:visualization}.
Without training our models to learn specific concepts, our model can locate and segment \texttt{Saturn V} as the lego rocket, \texttt{Oculus} as the VR headset, and \texttt{golden gate} as the bridge in corresponding images.
This demonstrates the strong potentials of open vocabulary semantic segmentation.

\subsubsection{Ambiguity of open vocabulary evaluation}

\begin{figure}[t]
    \begin{center}
    \includegraphics[width=\linewidth]{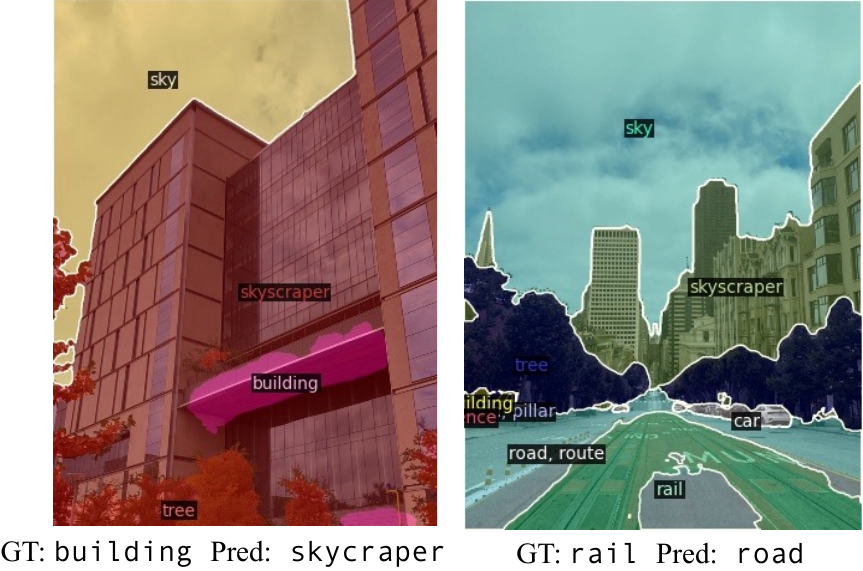}
    \vskip -1em
    \end{center}
    \vspace{-1em}
        \caption{Ambiguity of the class definition in open vocabulary segmentation evaluation.}
        \vspace{-1em}
    \label{fig:failure_cases}
\end{figure}

We show some ``failure'' predictions from the A-150 dataset in Figure~\ref{fig:failure_cases}.
For the left figure, the ground-truth category is ``building'' while our model predicts ``skyscrapers''.
The ``skyscrapers'' is a reasonable description, but the standard A-150 evaluation protocol will treat it as a wrong prediction.
A similar case happens in the right figure, the ground-truth ``rail'' is recognized as ``road''. 
This is caused by the fact that language defined categories are ambiguous and can overlap with each other.
Designing a better evaluation metric for open-vocabulary segmentation models is an important topic for our future research. Note that we use our own images, instead of ADE20K images in Figure ~\ref{fig:failure_cases}. But this phenomenon widely exists on ADE20K images.

\section{Conclusion}
This paper studies open-vocabulary semantic segmentation where the model segments an image by arbitrary categories described by texts.
We identify the performance bottleneck of current two-stage methods to be the pre-trained CLIP, since it doesn't perform well on masked images.
We propose to adapt CLIP for masked images.
To retain CLIP's open-vocabulary classification ability, we adapt CLIP with diverse mask-category pairs mined from image-caption dataset.
We further propose mask prompt tuning, a method can adapt CLIP without changing its original weights.
The proposed model is general and can do zero-shot segmentation on arbitrary datasets without dataset-specific adaption.
For the first time, we showopen-vocabulary generalist models can match the performance of supervised specialist models.

\section*{Acknowledgments}
We would like to thank Mengde Xu for setting up the baseline, Chenfeng Xu for helpful discussions.

Feng Liang and Diana Marculescu were partly supported in part by NSF CCF Grant No. 2107085 and NSF CSR Grant No. 1815780, as part of their affiliation with The University of Texas at Austin.

\section*{Ethics Statement}
We only use the public computer vision datasets (COCO, ADE20K, Pascal) and leverage the open-sourced vision-language models (CLIP) for our experiments. 
To the best of our knowledge, we do not foresee our approach as being inherently subject to concerns of discrimination / bias / fairness, inappropriate potential applications, impact, privacy and security issues, research integrity or research practice issues. However, the public datasets and pre-trained models may be subject to bias that may be inherited by models trained with our approach.

\section*{Reproducibility Statement}
Our code is reproducible and can be implemented based on the method description in Section~\ref{sec:method} as well as training details in Section~\ref{sec:setup}.

{\small
\bibliographystyle{ieee_fullname}
\bibliography{egbib}
}

\appendix
\section{Appendix}
\subsection{Crop with or without mask}
\label{appendix:mask_crop}
\begin{figure}[t]
    \begin{center}
    \includegraphics[width=0.7\linewidth]{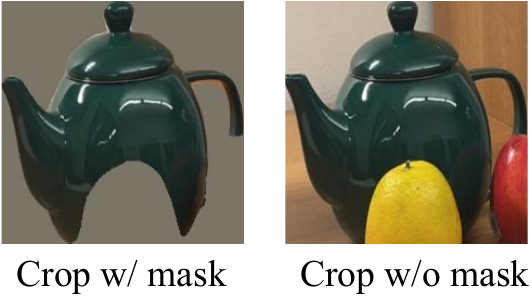}
    \end{center}
    \vspace{-1em}
        \caption{Crop without mask will introduce background pixels, making the prediction more difficult.}
    \label{fig:mask_crop}
\end{figure}

In the paper, we use the default crop with mask (see left of Figure~\ref{fig:mask_crop}).
We also try the direct crop without mask  (see right of Figure~\ref{fig:mask_crop}). 
Following the bottleneck analysis in the introduction, we feed the unmasked crops a pre-trained CLIP for classification. 
This experiment gives a 13.8\% mIoU, which is -6.3\% worse than using the masked crops.
We hypothesize that the crop with mask introduces many background pixels, making the prediction more difficult.
For the example in the right of Figure~\ref{fig:mask_crop}, the ``orange'' will also be an appropriate category for the unmasked crop.

We note that ZegFormer~\cite{ding2022zegformer} has also done an ablation study about different strategies to obtain the final crop.
We reach a similar conclusion.

\subsection{Text templates}
We use the text templates from ViLD~\cite{gu2021vild}.
For each category, we used multiple templates to generate the text embeddings then ensemble these embeddings by a simple average.
Text templates are shown as below: 

\begin{verbatim}
'a photo of a {}.',
'This is a photo of a {}',
'There is a {} in the scene',
'There is the {} in the scene',
'a photo of a {} in the scene',
'a photo of a small {}.',
'a photo of a medium {}.',
'a photo of a large {}.',
'This is a photo of a small {}.',
'This is a photo of a medium {}.',
'This is a photo of a large {}.',
'There is a small {} in the scene.',
'There is a medium {} in the scene.',
'There is a large {} in the scene.',
\end{verbatim}

\subsection{Class prediction ensemble weight}
\label{appendix:ensemble_weight}

\begin{table}[t]
\caption{The effects of class prediction ensemble. The baseline and our OVSeg model are Swin-Base + Vit-L. We report the mIoU on A-150.
}
\small
\vspace{0.2cm}
\centering
\begin{tabular}{cccc}
\toprule
         & MaskFormer only & CLIP only & Ensemble \\
\midrule \midrule
baseline & \textbf{19.6}            & 14.3      & 21.8     \\
\ours (Ours)    & \textbf{19.6}            & \textbf{25.1}      & \textbf{29.6}     \\
\bottomrule
\end{tabular}
\label{tab:class_ensemble}
\end{table}

We set $\lambda=0.7$ for A-150 and A-847, $\lambda=0.6$ for PAS-20, PC-59 and PC-459.
We further detail the effects of ensemble on A-150 in Table.~\ref{tab:class_ensemble}.
MaskFormer only or CLIP only denotes the use of the class prediction of MaskFormer or CLIP only. 
Compared with the baseline, we adapt the CLIP to masked images, leading to a much better CLIP only performance. 
We also notice ensemble is essential for good performance.

\subsection{Training hyperparams of R101c model}
\label{appendix:r101c_vitb}
Our small model is MaskFormer R101c with CLIP ViT-B/16.
For MaskFormer training, the backbone weights are initialized from an ImageNet-1K pre-trained model. We use AdamW optimizer with the poly learning rate schedule. 
The initial learning rate and weight decay are set to $2 \cdot 10^{-4}$ and $10^{-4}$, respectively. 
We also use a learning rate multiplier $0.1$ on the backbone.
We use a crop size of $512 \times 512$, a batch size of 32 and train the model for 120K iterations.
For data augmentations and other hyper-parameters, we follow the setting of ~\cite{cheng2021maskformer}.
For adapting CLIP ViT-B/16 model, we basically follow the hyperparameters of finetuning ViT-L/16 except we use a larger batch size 1024.

\subsection{More ablation studies on mask prompt tuning}
\label{appendix:details_of_mask_prompt_tuning}

\begin{table}[t]
        \caption{Ablation on combining mask prompt tuning (MPT) and fine-tuning (FT). 
        FT -\textgreater MPT indicates first FT and then MPT, and vice versa.
        FT + MPT sim. means optimizing prompts and CLIP simultaneously.
        }
        \label{tab:simultaneous_MPT_ft}
        \centering
        \vspace{0.5em}
        \begin{tabular}{l|cc}
        \toprule
           combination    & A-847 & A-150 \\
        \midrule \midrule
        FT -\textgreater MPT (default) &  \textbf{9.0}    & \textbf{29.6}  \\
        MPT -\textgreater FT &  8.5 \color{red}{(-0.5)} &  28.1 \color{red}{(-1.5)}   \\
        FT + MPT sim.        &  8.8 \color{red}{(-0.2)}    & 29.0 \color{red}{(-0.6)}\\
        \bottomrule
        \end{tabular}
\end{table}

We explore two other ways to combine mask prompt tuning (MPT) and fine-tuning (FT) as in Table~\ref{tab:simultaneous_MPT_ft}. 
Our default setting (FT -\textgreater MPT) is first doing FT and then applying MPT to the already fine-tuned model.
We don't change the weights of fine-tuned CLIP.
The other option is first doing MPT and then doing FT with fixed mask prompts (MPT -\textgreater FT).
This combination produces poor results (-1.5\% drop on A-150).
We conjecture mask prompts learned with original CLIP provide a bad prior when we fune-tune the entire CLIP model.
We also explore learning mask prompts and fine-tune CLIP weight \emph{simultaneously} (FT + MPT sim.).
This doesn't bring favorable results either.

\begin{table}[t]
        \caption{Ablation on prompt depth. We test with and without fully fine-tuned (FT) model.
        }
        \label{tab:prompt_depth}
        \centering
        \vspace{0.5em}
        \begin{tabular}{l|cc}
        \toprule
        \multirow{2}{*}{prompt depth} & \multicolumn{2}{c}{A-150}                              \\
        \cmidrule(lr){2-3}
                                      & \multicolumn{1}{c}{w/o FT} & \multicolumn{1}{c}{w/ FT} \\
        \midrule \midrule
        1                             &  25.7                           &  29.3                         \\
        3  (default)                     &  26.5                  &   \textbf{29.6}               \\
        6                             &   \textbf{26.8}                     &  29.4                         \\
         12                             &   \textbf{26.8}                 &  29.3                         \\

        \bottomrule
        \end{tabular}
\end{table}

We further ablate the effects of prompt depth in Table~\ref{tab:prompt_depth}. 
The depth can be selected from $\{1,3,6,12\}$. 
We use two different scenarios: without fine-tuning (w/o FT) for mask prompt tuning only, with fine-tuning (w/ FT) for applying mask prompt tuning over a already fine-tuned model. 
For w/o FT case, one layer prompt can bring significant improvement, \eg, from baseline's 21.8\% to 25.7\%. 
Deeper prompts result in better performance, because more parameters are introduced with more prompts.
Interestingly, deeper prompts (going from 3 to 12) don't bring further improvement for w/ FT case.
We choose prompt depth as 3 for default setting.

\subsection{Compare masked prompt tuning (MPT) to Deep Visual Prompt Tuning (VPT)~\cite{jia2022vpt}}

We compared our MPT to VPT~\cite{jia2022vpt}. 
With the Swin-Base + ViT-L/14 baseline, we added 50 learnable tokens to the image input tokens. VPT used "deep prompts" with depth 6, resulting in 25.5\% mIoU on A-150, which is 1.0\% worse than MPT (case (a) in Table 3). This could be due to the use of masked prompts in MPT, which prevent zero masked tokens and mitigate domain distribution shifts in the CLIP model. Additionally, MPT requires no additional computation, while VPT requires 40\% more computation to process the extra tokens.  We plan to include this ablation study in our final draft.

\begin{table}[t]
\centering
\caption{Comparison between different prompt tuning methods.}
\begin{tabular}{cccc}
\toprule
Method    & baseline & MPT (ours) & VPT \\ \midrule \midrule
mIoU on A-150 & 21.8  & \textbf{26.5} & 25.5       \\ \bottomrule
\end{tabular}
\end{table}

\subsection{Combine training pairs from COCO-stuff and COCO-Caption pseudo segments.}

We combined GT COCO-stuff annotations (case (1) in Tab.2) with caption pseudo-labeled annotations (case (3) in Tab.2), resulting in 1.4M pairs with 12K nouns in Table~\ref{tab:source}. However it underperformed compared to using only pseudo-labeled annotations (26.7\% mIoU {\it vs.} 28.8\% mIoU on A-150). We believe the class distribution was dominated by the GT COCO-stuff annotations and resulted in overfitting. Future work could explore a more balanced data selection ({\it e.g.} 10\% GT + 90\% pseudo-labeled annotations) to potentially improve performance.

\begin{table}[t]
\centering
\caption{The source of mask-category training pairs.}
\begin{tabular}{cccc}
\toprule
Training pairs    & Stuff & Cap. & Stuff + Cap. \\ \midrule \midrule
mIoU on A-150 & 23.0  & \textbf{28.8} & 26.7       \\ 
\bottomrule
\end{tabular}
\label{tab:source}
\end{table}

\subsection{Class-wise IoU over seen and unseen categories.}
\label{appendix:class_iou}
\begin{figure}[h]
    \centering
	\includegraphics[width=2.2\columnwidth,angle=270]{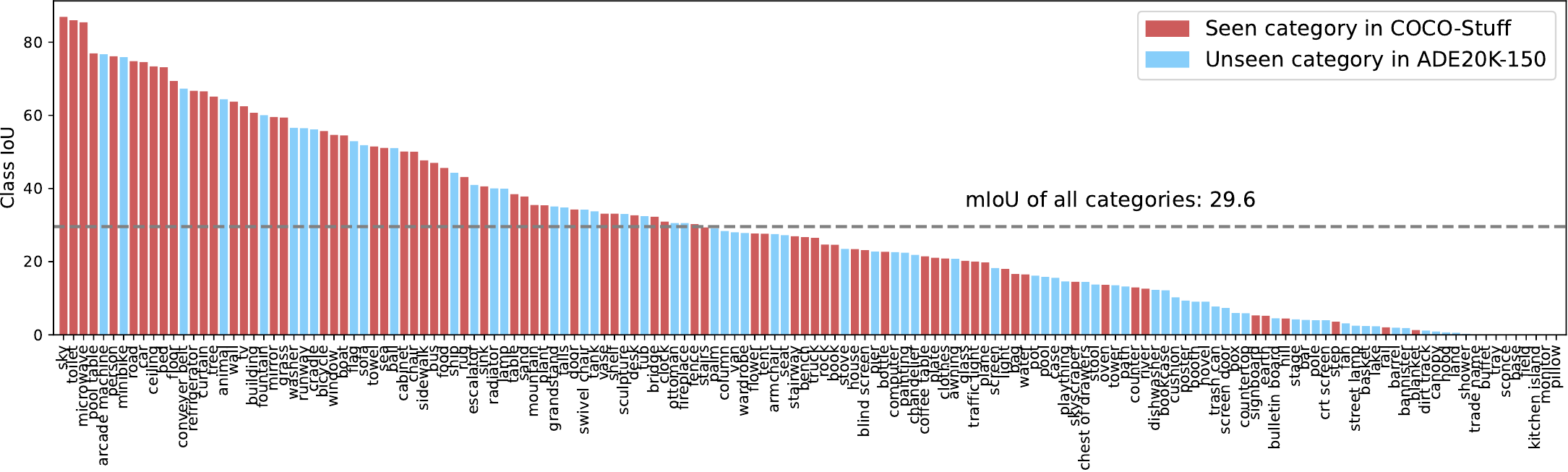}   
	\caption{Class IoU on all 150 categories in ADE20K (model trained on COCO). It is expected the model performs better on seen categories in training set.}
	\label{fig:class_iou}
\end{figure}

We detail the class IoU on all 150 categories in ADE20K-150 (model trained on COCO) in Figure~\ref{fig:class_iou}, and we annotated seen \textit{vs.} unseen classes and their IoUs. Seen categories mean there are \textit{similar} categories in COCO-stuff training set. Unseen categories denote the novel categories in ADE20K. 
The average IoU of seen and unseen categories are 37.6\% and 21.9\%, respectively, showing that our model performs better on seen categories. This is also observed in other open vocabulary segmentation work, such as ~\cite{ding2022zegformer}.

\subsection{Inference speed discussions}

We followed the two-stage framework of SimBaseline~\cite{xu2021simple} with a focus on accuracy improvement. Our study also evaluated the inference time of MaskFormer and CLIP region classification. For our OVSeg model (Swin-Base + ViT-L), the inference time of MaskFormer and CLIP is roughly 0.2s and 0.6s, respectively, per image on an NVIDIA A5000 GPU. We acknowledge that processing hundreds of regions with CLIP is time-intensive and understand that improving the efficiency of two-stage frameworks is a crucial area of research. It is out of the scope of this work and we plan to address this challenge in future work.

\end{document}